\documentclass[letterpaper, 10 pt, conference]{ieeeconf}  % Comment this line out if you need a4paper
\usepackage{xcolor}
\usepackage{cite} % IEEE 要求的引用包

\usepackage{booktabs}
\usepackage{multirow}
\usepackage{placeins}
\FloatBarrier

\usepackage{amsmath}
\usepackage{amssymb}

\usepackage{graphicx}
\usepackage{multirow}

\usepackage{array}
\usepackage{caption}
\usepackage{float}
\usepackage{tabularx} 
\usepackage{makecell}

\usepackage{colortbl}
\usepackage{booktabs}

\usepackage{url} % 使用 \url 命令

\definecolor{myblue}{RGB}{0,102,204}
\definecolor{darkgreen}{rgb}{0.0, 0.39, 0.0}  % 深绿色
\definecolor{darkred}{rgb}{0.55, 0.0, 0.0}    % 深红色

\makeatletter
\let\NAT@parse\undefined
\makeatother

\usepackage[colorlinks=true, linkcolor=myblue, citecolor=myblue, urlcolor=myblue]{hyperref}

\IEEEoverridecommandlockouts                              % This command is only needed if 
\title{\LARGE \bf
RoboMatch: A Unified Mobile-Manipulation Teleoperation Platform with Auto-Matching Network Architecture for Long-Horizon Tasks
}

\author{Hanyu Liu$^{*,1}$, Yunsheng Ma$^{*,1}$, Jiaxin Huang$^{1}$, Keqiang Ren$^{1}$, Jiayi Wen$^{1}$,  Yilin Zheng$^{1}$, \\
Haoru Luan$^{1}$, Baishu Wan$^{1}$, Pan Li$^{1}$, Jiejun Hou$^{1}$, Zhihua Wang$^{1}$, Zhigong Song$^{1}$ % <-this % stops a space
\thanks{$^{*}$These authors contributed equally to this work.}% <-this % stops a space
\thanks{Corresponding author: Zhigong Song ({song\_jnu}@jiangnan.edu.cn).} %
\thanks{
$^{1}$Jiangsu Key Laboratory of Advanced Food Manufacturing Equipment and Technology, School of Mechanical Engineering, Jiangnan University, Wuxi 214122, China. This work was supported by the Wuxi Taihu Lake Talent Plan.
}
}

\begin{document}

% \begin{CJK*}{UTF8}{gbsn}
\maketitle
\thispagestyle{empty}
\pagestyle{empty}

%%%%%%%%%%%%%%%%%%%%%%%%%%%%%%%%%%%%%%%%%%%%%%%%%%%%%%%%%%%%%%%%%%%%%%%%%%%%%%%%

\begin{abstract}

This paper presents RoboMatch, a novel unified teleoperation platform for mobile manipulation with an auto-matching network architecture, designed to tackle long-horizon tasks in dynamic environments. Our system enhances teleoperation performance, data collection efficiency, task accuracy, and operational stability. The core of RoboMatch is a cockpit-style control interface that enables synchronous operation of the mobile base and dual arms, significantly improving control precision and data collection. Moreover, we introduce the Proprioceptive-Visual Enhanced Diffusion Policy (PVE-DP), which leverages Discrete Wavelet Transform (DWT) for multi-scale visual feature extraction and integrates high-precision IMUs at the end-effector to enrich proprioceptive feedback, substantially boosting fine manipulation performance. Furthermore, we propose an Auto-Matching Network (AMN) architecture that decomposes long-horizon tasks into logical sequences and dynamically assigns lightweight pre-trained models for distributed inference. Experimental results demonstrate that our approach improves data collection efficiency by over 20\%, increases task success rates by 20–30\% with PVE-DP, and enhances long-horizon inference performance by approximately 40\% with AMN, offering a robust solution for complex manipulation tasks. Project website: \href{https://robomatch.github.io}{\textcolor{myblue}{\texttt{https://robomatch.github.io}}}

\end{abstract}

%%%%%%%%%%%%%%%%%%%%%%%%%%%%%%%%%%%%%%%%%%%%%%%%%%%%%%%%%%%%%%%%%%%%%%%%%%%%%%%%
\section{INTRODUCTION}
Imitation learning from human demonstrations offers an efficient way for robots to acquire skills, allowing humans to teach various tasks. Mobile teleoperation platforms are ideal for complex whole-body coordination tasks like opening doors or moving objects. Current systems mainly follow two approaches: decoupled platforms~\cite{RN17} require distinct operations for mobility and manipulation. While simplifying individual control, this divided approach may cause efficiency loss and synchronization issues. Integrated platforms face either usability limitations (e.g., Mobile ALOHA's push-pull control~\cite{RN2}) or flexibility constraints (e.g., HOMIE's fixed cockpit~\cite{RN44}). These limitations hinder high-quality demonstration collection and restrict imitation learning's effectiveness in complex tasks. 

% Thus, developing convenient and flexible teleoperation platforms has become crucial for advancing robot imitation learning.

% 为了减少人力成本，让一个人能同时轻松完成双臂和移动的数据采集，我们开发了低成本全新移动机器人数据采集平台—RoboMatch，包括可移动式主动端控制支架和IMU腰带（适脚器）运动控制系统，为动态环境下的移动机器人操作提供了全新的数据采集与交互范式。惯性测量单元（IMU）在机械臂控制中已经被广泛运用，如UMI。在我们的工作中，我们创新地将IMU与底盘控制相结合，搭建了IMU腰带（适脚器）运动控制系统。通过该设计，在移动机器人数据采集过程中，可以一个人轻松完成数据采集工作，减少人力成本的同时，提高了任务完成质量（因为两个人操作没有一个人操作完成效果好）。

Imitation learning has gained significant attention in robotics due to its practical applicability. Strategies like ACT~\cite{RN4} and Diffusion Policy~\cite{RN5} enable autonomous decision-making through end-to-end learning from demonstrations and environmental perception. Such perception includes visual~\cite{RN37} and proprioceptive sensing~\cite{RN39}. While visual representations encode scene information, learning robust visual features for robotics remains challenging. Attention mechanisms like the Efficient Multi-scale Attention Module (EMA)~\cite{RN43} enhance visual perception but focus only on spatial features, ignoring frequency-domain information. Proprioceptive accuracy critically affects motion control precision. During delicate tasks like assembly or grasping, insufficient data and error accumulation often cause end-effector pose deviations, leading to failure. 

% Improving proprioception and reducing error propagation are thus key challenges in embodied intelligence.

% 在此基础上，我们设计了本体-视觉感知增强扩散策略（PVE-DP Policy），在Diffusion Policy的视觉骨干网络中嵌入离散小波变换以增强视觉表征。在FE-EMA中，论证了频域信息对机器人操作任务的重要性。因此，我们结合其思想，将其应用在扩散模型中，发现频域数据能显著地提高Diffusion Policy的动作生成质量。

% 在模仿学习中，机器人操作精度不够理想，这可能是本体感知不够，目前大部分工作的本体感知仅停留在关节角度，以及少部分工作加入了触觉感知。为此，我们在机械臂末端集成惯性测量单元（IMU）。

% IMU是一种多轴运动传感器，通常包含三轴加速度计、三轴陀螺仪和磁力计，能够实时测量末端执行器的加速度、角速度和姿态信息。通过高频率的数据采集，IMU可有效补偿机械臂运动过程中的动态误差，并为后续的状态估计和运动控制提供精确的惯性参考。此外，IMU的轻量化和小型化设计使其易于集成到机械臂末端，而不会显著增加系统负载或影响运动灵活性。我们实时检测任务中夹爪的末端四元数数据，并将其融入算法训练。这么做，显著提升了机器人进行精细操作时的本体感知能力~\cite{RN1} ，并在实际任务测试过程中取得了很好的效果。

As robotic intelligence advances, task planning, as a critical bridge connecting high-level instructions with low-level execution, faces growing challenges. Traditional rule-based methods often fall short in long-horizon, multi-step tasks within open environments. Emerging large language models (LLMs)~\cite{RN14} offer a promising alternative by parsing abstract instructions into executable task sequences through natural language understanding and logical reasoning. When integrated with visual-language models (VLMs)~\cite{RN42, RN55}, robots can transcend simple perception to achieve complex task reasoning. For instance, ReKep~\cite{RN56} leverages VLMs to generate physically grounded cost functions over structured 3D representations, enabling precise, constrained strategy execution. Breakthrough end-to-end vision-language-action (VLA)~\cite{RN53, RN54, RN34} architectures further unify perception, decision-making, and execution in a single model. However, critical challenges remain: constraint satisfaction struggles with complex multi-step tasks; VLAs exhibit limited long-horizon reasoning and hallucinations~\cite{RN60}; while substantial network architectures cause latency and error accumulation, hindering real-world deployment~\cite{RN59}.

% Overcoming these bottlenecks is essential to enhance the reliability and adaptability of robotic task planning.

Therefore, we propose \textbf{RoboMatch}, a unified mobile-manipulation teleoperation platform with a mecha aesthetic and auto-matching network architecture for long-horizon tasks. The main contributions presented in our paper are as follows:

1) \textbf{Unified Mobile-Manipulation Teleoperation Platform:} We propose RoboMatch, a mecha-inspired platform integrating unified mobility-manipulation control, exoskeleton operation, VR-assisted perception, \textit{etc.}, for high-quality demonstration collection.

% \begin{figure*}[htbp]
%     \centering
%     \includegraphics[width=\textwidth]{Pictures/2true_6.png} % 适应整个页面宽度
%     \captionsetup{singlelinecheck=off, justification=raggedright}
%     \caption{Overview of RoboMatch.RoboMatch实现操作座舱与机器人本体深度融合，支持座舱内精准操控。遥操作阶段，操作者通过VR多视角观测作业画面；座舱集成操作系统便于调试，两侧双主动臂实现主从臂精确映射；底盘左右脚踏板基于IMU的Pitch轴角度变化，分别控制进退与转向。推理阶段，凭借高集成设计，可在多样场景与任务中高效完成数据采集及推理。}
%     \label{PVE-DP}
% \end{figure*}

\begin{figure*}[htbp]
    \centering
    \includegraphics[width=\textwidth]{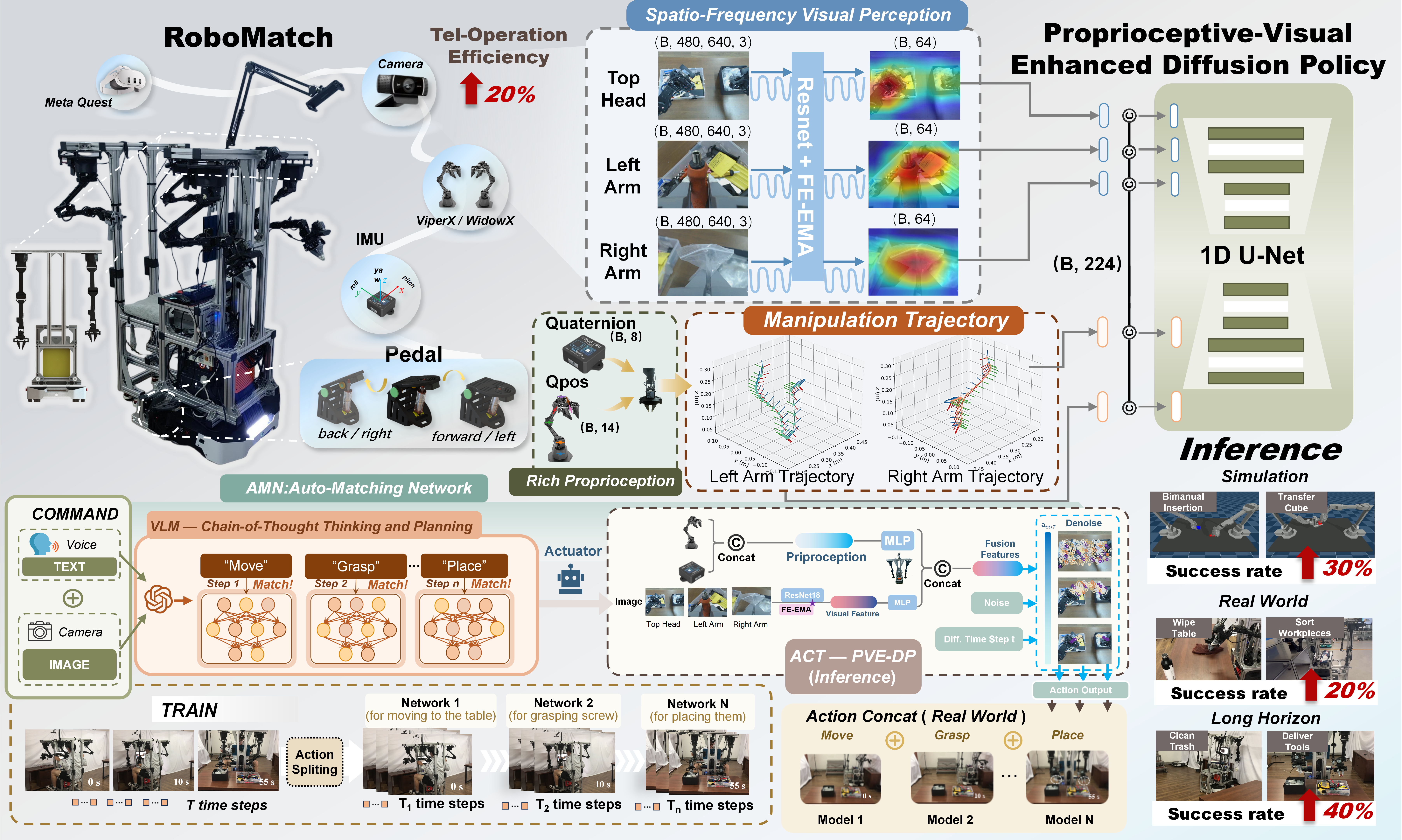} % 适应整个页面宽度
    \captionsetup{singlelinecheck=off, justification=raggedright}
    \caption{Overview of the RoboMatch framework. This figure integrates three core components: (1) \textbf{RoboMatch}, a unified mobile-manipulation teleoperation platform that matches the robot base with the operation platform to achieve high-precision master-slave control and immersive observation; (2) \textbf{PVE-DP}, a policy combining spatio-frequency visual enhancement and rich proprioception to improve fine manipulation accuracy; (3) \textbf{AMN}, an architecture that combines the semantic parsing capabilities of VLMs with the execution advantages of specialized small policy networks for long-horizon execution, enabling chain-of-thought reasoning for complex task decomposition and adaptive operation.}
    \label{Overview_RoboMatch}
\end{figure*}

2) \textbf{Proprioceptive-Visual Enhanced Diffusion Policy:} We design the PVE-DP framework, introducing enhancements in two key areas:
\begin{itemize}
\item We integrate the \text{FE-EMA} visual module (Frequency Enhanced EMA) into the \text{Diffusion Policy} visual backbone network to construct spatio-frequency-fused visual representations.
\item We mount IMU sensors on both end-effectors to accurately perceive rotational states by capturing real-time quaternion data from each arm. This information is fused with joint angles to form a comprehensive proprioceptive representation.
\end{itemize}

3) \textbf{Auto-Matching Network (AMN):} This work introduces an AMN framework that combines VLM-based global planning with efficient compact policy execution. Through chain-of-thought reasoning, it decomposes complex tasks into hierarchical subtasks and automatically assigns them to pre-trained lightweight policy networks.

Experimental results show RoboMatch achieves 20\% higher data collection efficiency. PVE-DP improves task success rates by 30\% in simulation and 20\% in real-world ALOHA tasks, AMN improves inference performance by around 40\% in long-horizon tasks. These advancements provide an effective solution for robotic data collection, precise manipulation, and long-horizon task execution.

\section{RELATED WORK}

\subsection{Teleoperation for Mobile Manipulation}

Robot Teleoperation~\cite{RN11,RN12} enables real-time remote control via intermediary systems and is widely used in industrial, medical, and collaborative scenarios. However, mobile manipulation research remains nascent, especially in achieving efficient mobile base-arm coordination. While early systems adopted master-slave architectures~\cite{RN18}, recent advances like Mobile ALOHA~\cite{RN2} and HOMIE~\cite{RN44} integrate mobility and manipulation. Some approaches leverage high-end hardware (e.g., motion capture~\cite{RN23}, exoskeletons~\cite{RN44}) or adapt static interfaces (e.g., VR~\cite{RN20}, keyboard~\cite{RN22}) to mobile platforms. Yet, as shown in Table~\ref{tab:teleoperation_comparison}, most solutions still lack fine multi-DoF control, task generalization, operational integration, and long-term deployment capabilities, limiting their applicability in complex mobile manipulation tasks requiring high precision and integration.

\begin{table*}[htbp]
\centering
\caption{Comparison of Existing Robot Teleoperation Systems}
\label{tab:teleoperation_comparison}
\resizebox{\textwidth}{!}{
\begin{tabular}{l c c c c c c c}
\toprule
\makecell{\textbf{Teleop System}} & 
\makecell{\textbf{Modality}} & 
\makecell{\textbf{Action} \\ \textbf{Space}} & 
\makecell{\textbf{Cockpit}} & 
\makecell{\textbf{Bimanual}} & 
\makecell{\textbf{Exo-} \\ \textbf{Teleop}} & 
\makecell{\textbf{Whole-Body} \\ \textbf{Teleop}} & 
\makecell{\textbf{Wild}} \\
\midrule
Mobile ALOHA~\cite{RN2} & Puppeteer & Joint Pos. / Base Vel. &
\textcolor{darkred}{$\times$} & \textcolor{darkgreen}{$\checkmark$} & \textcolor{darkred}{$\times$} &
\textcolor{darkgreen}{$\checkmark$} & \textcolor{darkgreen}{$\checkmark$} \\
AirExo~\cite{RN49} & Puppeteer & Joint Pos. &
\textcolor{darkred}{$\times$} & \textcolor{darkgreen}{$\checkmark$} & \textcolor{darkgreen}{$\checkmark$} &
\textcolor{darkred}{$\times$} & \textcolor{darkgreen}{$\checkmark$} \\
MoMaRT~\cite{RN17} & Phone & EE Pose / Base Vel. &
\textcolor{darkred}{$\times$} & \textcolor{darkred}{$\times$} & \textcolor{darkred}{$\times$} & \textcolor{darkgreen}{$\checkmark$} &\textcolor{darkred}{$\times$} \\
Dexcap~\cite{RN50} & Vision Retarget & EE Pose &
\textcolor{darkred}{$\times$} & \textcolor{darkgreen}{$\checkmark$} & \textcolor{darkgreen}{$\checkmark$} &
\textcolor{darkred}{$\times$} & \textcolor{darkred}{$\times$} \\
MOMA-Force~\cite{RN46} & Kineshetic & EE Pose and Wrench &
\textcolor{darkred}{$\times$} & \textcolor{darkgreen}{$\checkmark$} & \textcolor{darkred}{$\times$} &
\textcolor{darkred}{$\times$} & \textcolor{darkred}{$\times$} \\
UMI~\cite{RN51} & Kineshetic & EE Pose &
\textcolor{darkred}{$\times$} & \textcolor{darkred}{$\times$} & \textcolor{darkred}{$\times$} &
\textcolor{darkred}{$\times$} & \textcolor{darkgreen}{$\checkmark$} \\
HOMIE~\cite{RN44} & Puppeteer / Pedal & Joint Pos. / Gait &
Fixed & \textcolor{darkgreen}{$\checkmark$} & \textcolor{darkgreen}{$\checkmark$} &
\textcolor{darkgreen}{$\checkmark$} & \textcolor{darkred}{$\times$} \\
\midrule
\textbf{RoboMatch (ours)} & VR / Puppeteer / Pedal / Multi-View & Joint Pos. / Base Vel. &
Followed & \textcolor{darkgreen}{$\checkmark$} & \textcolor{darkgreen}{$\checkmark$} &
\textcolor{darkgreen}{$\checkmark$} & \textcolor{darkgreen}{$\checkmark$} \\
\bottomrule
\end{tabular}
}
\end{table*}

% In contrast, RoboMatch deeply integrates the robot and control cockpit, delivering unified mobility-manipulation control, high-precision master-slave motion mapping, ergonomic exoskeleton operation, and VR-enhanced visual perception. The system balances precision, usability, and integration, offering an efficient and extensible teleoperation paradigm for complex mobile manipulation.

\subsection{Imitation Learning}

Imitation learning has become a key approach in robotics~\cite{RN26,RN28}, enabling robots to acquire manipulation strategies for complex tasks from expert demonstrations~\cite{RN29,RN30}. Among these methods, behavior cloning (BC) serves as a foundational technique that directly maps observations to actions~\cite{RN31}, offering strong transferability and proving particularly suitable for policy learning based on teleoperation data. Recent advances integrate imitation learning with deep learning frameworks, adopting network strategies such as Transformer~\cite{RN4,RN32} and Diffusion~\cite{RN5} for autonomous decision-making in complex continuous tasks. Notably, diffusion models have demonstrated remarkable generalization in handling intricate robotic tasks and are widely adopted in VLA large models~\cite{RN34,RN35,RN36}. 

Despite their strong task adaptability, diffusion models still exhibit noticeable limitations in motion generation precision. Cumulative errors during inference, coupled with insufficient visual perception and proprioception, significantly impede precise decision-making and efficient robot control~\cite{RN61}. 
% 模仿学习已经成为机器人领域的关键方法之一~\cite{RN26,RN27,RN28}，通过专家演示帮助机器人学习复杂任务中的操作策略~\cite{RN29,RN30}。其中，行为克隆（BC）作为基础方法，将观测直接映射为动作~\cite{RN31}，具有良好的可迁移性，特别适用于基于远程操作数据进行策略学习。近年来，模仿学习与深度学习框架的结合取得了显著进展，尤其是在复杂连续任务的自主决策中，采用了Transformer~\cite{RN31,RN4,RN32}和扩散模型（Diffusion）~\cite{RN5}等网络策略。特别是扩散策略，在处理机器人复杂任务时表现出卓越的泛化能力，因此被广泛应用于Vision-Language-Action（VLA）大模型中~\cite{RN34,RN35,RN36}。尽管扩散模型在任务适应性上展现了强大能力，但在动作生成的精度方面仍存在显著不足。尤其是在精细操作任务（如叠衣服）中，大多数现有算法的本体感知仅依赖于关节角度，且视觉模块通常仅使用简单的ResNet-18结构。然而，推理过程中存在的累计误差，以及视觉感知~\cite{RN37,RN38}与本体感知~\cite{RN39}信息的不充分，严重影响了机器人的精确决策与高效控制。

% In contrast, our proposed PVE-DP enhances both proprioceptive and visual perception capabilities. For proprioception, we incorporate end-effector quaternion data (Quat), enabling real-time monitoring of 3D rotational states. For visual perception, Discrete Wavelet Transform (DWT) is employed to extract frequency-domain features from temporal images, achieving time-frequency fusion. Experimental results demonstrate that our method outperforms state-of-the-art (SOTA) approaches across four simulated and real-world manipulation tasks.

\subsection{Long-Horizon Inference in Robotics}
% Recent studies show that LLM-based planning agents offer substantial advantages in long-horizon task decomposition and execution~\cite{RN14,RN40}. 

The decomposition and execution of long-horizon tasks have long been a core focus in robotics~\cite{RN40}, with recent studies demonstrating that LLM-based planning agents offer substantial advantages in this domain~\cite{RN14}. The rapid development of large language models has made natural language instruction prompting a major trend in robot task planning. Recently, VLMs have demonstrated broad potential, with research mainly dividing into two technical pathways. In the first pathway, some approaches employ VLMs as task planners or formulate them as constraint satisfaction problems~\cite{RN42}. For instance, the ReKep system~\cite{RN56} uses VLMs to generate numerical cost functions defined over 3D environmental keypoints, optimizing relational constraints across the robot-object system to achieve multi-stage task execution. The second pathway explores integrated VLA models that combine VLMs with action heads, providing a feasible framework for end-to-end robot learning, as exemplified by OpenVLA~\cite{RN53} and $\pi$0.5~\cite{RN54}. 

However, these methods still face notable limitations: constraint-based planners perform well in simple pick-and-place tasks but degrade significantly in multi-step scenarios, while VLA models suffer from inadequate long-horizon inference, frequent hallucinations, and slow inference speeds due to their large architectures~\cite{RN59,RN60}. These issues amplify error accumulation and result in substantially lower task success rates. 

% In contrast, our proposed AMN framework integrates VLMs’ task planning strengths with the execution efficiency of compact policy networks. It decomposes complex tasks into structured subtasks via chain-of-thought reasoning and automatically assigns each to a pre-trained lightweight network for sequential execution. This hybrid design preserves high-level semantic understanding from VLMs while enhancing subtask execution reliability through specialized networks. It significantly improves long-horizon success rates, system stability, and complex task handling.

\section{METHOD}

\begin{figure*}[htbp]
    \centering
    \includegraphics[width=\textwidth]{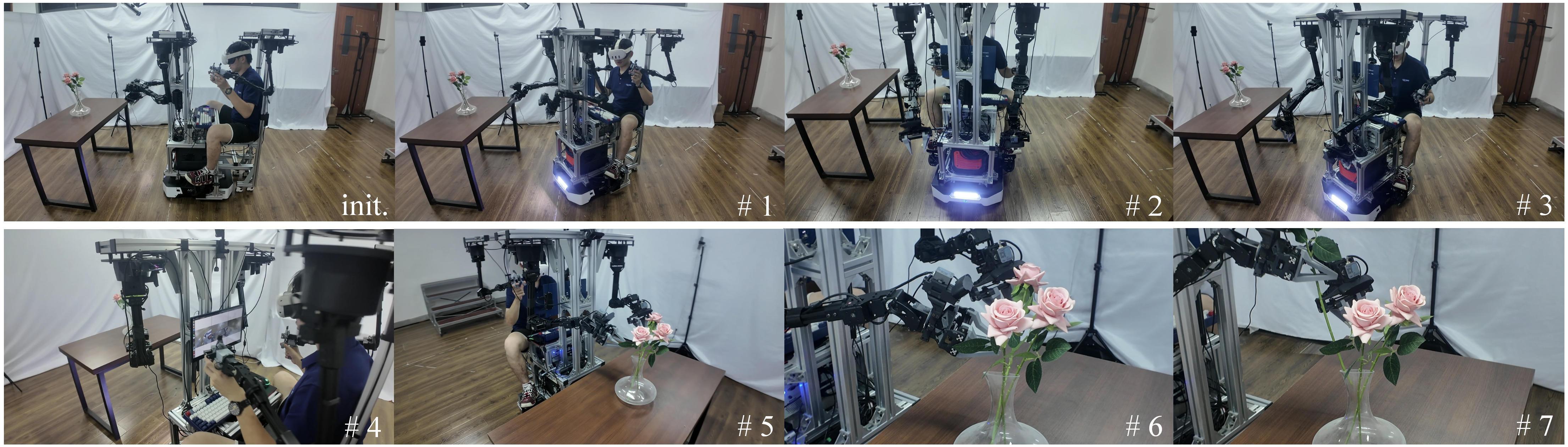} % 适应整个页面宽度
    \captionsetup{singlelinecheck=off, justification=raggedright}
    \caption{RoboMatch teleoperation demonstration.}
    \label{Tel_Operation_real}
\end{figure*}

As shown in \hyperref[Overview_RoboMatch]{Fig.~\ref{Overview_RoboMatch}}, RoboMatch is a unified mobile-manipulation framework for data collection and policy learning, capable of directly capturing human demonstrations in real-world environments and converting them into deployable robot policies. The system is designed with the following objectives:

\textbf{Collaborative:} RoboMatch integrates the robot and cockpit for seamless human-controlled mobile manipulation, enabling a single operator to perform grasping and mobility tasks effortlessly.
    
\textbf{Precise:} Our system captures end-effector (EE) rotational states and frequency-domain information from visual data, enhancing manipulation accuracy.

\textbf{Long-Horizon:} Our framework decomposes long horizon tasks into subtask actions through a chain-of-thought reasoning paradigm, dynamically assembling specialized neural networks to accomplish complex long-horizon robotic operations.
    
The following sections elaborate on how we achieve the above objectives through hardware design and algorithmic strategy design.

\subsection{RoboMatch Hardware Design and Manipulation} 

Our designed humanoid robot is named \textbf{RoboMatch}. The system consists of two main components: the robot body and the control cockpit. The robot body comprises two 7-DOF slave manipulators (ViperX-300), an indoor differential-drive mobile base (Tracer-2.0), three Logitech RGB cameras (C922x webcams), and two end-effector IMUs. The cockpit is equipped with two 7-DOF master arms (WidowX-250), foot pedals for base motion control, a display, keyboard, mouse, and an industrial computer. The human operator wears a VR headset (Meta Quest-3) for real-time task monitoring through multi-view visual feedback.

In terms of design, we adopted an inverted master-slave manipulator layout to form an exoskeleton-based teleoperation configuration, enhancing operational intuitiveness by aligning with the operator’s natural motion mapping. Dedicated foot pedals with embedded IMUs were developed for feet-controlled base movement. Each pedal measures rotation around the Pitch axis, mapping the angle to velocity commands: the left pedal controls forward/backward motion with speed proportional to tilt, while the right pedal enables precise left/right rotation using the same mapping logic.

As shown in \hyperref[Tel_Operation_real]{Fig.~\ref{Tel_Operation_real}}, RoboMatch enables the operator to effortlessly perform both mobility and fine manipulation through the joint mapping of the master-slave manipulators and multi-view visual feedback. The system enables coordinated upper-lower limb movement in a unified human-robot drive, significantly improving ergonomic adaptation and immersive control. The highly integrated hardware-software architecture not only enhances data collection efficiency but also provides high-fidelity and highly consistent human demonstration data for imitation learning and teleoperation tasks.

\begin{figure*}[htbp]
    \centering
    \includegraphics[width=\textwidth]{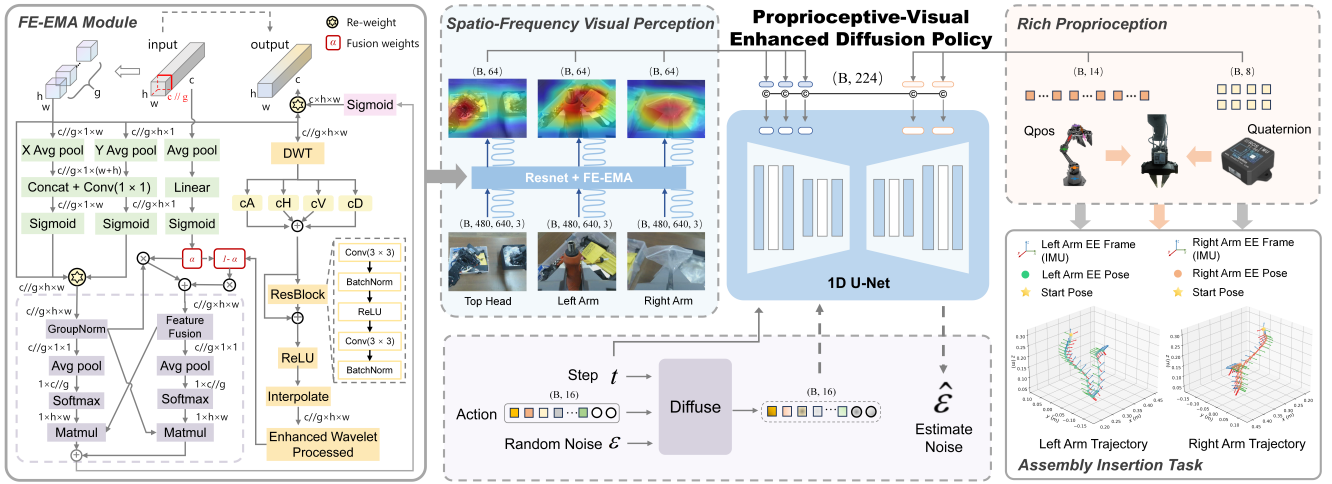} % 适应整个页面宽度
    \captionsetup{singlelinecheck=off, justification=raggedright}
    \caption{Overall network architecture of PVE-DP, which consists of three core modules: (1) \textbf{FE-EMA Module} enhances robotic visual perception by integrating spatio-frequency features from visual data; (2) \textbf{Rich Proprioception} strengthens the robot's self-state awareness through concatenating arm joint positions (Qpos) and end-effector quaternion (Quat) from IMU sensing; (3) \textbf{DP U-Net} takes the enhanced visual-proprioceptive observations as conditions and predicts fine-grained action noise to achieve precise manipulation during robot inference.}
    \label{PVE-DP}
\end{figure*}

\subsection{The Model Framework of PVE-DP}

% 在机器人模仿学习领域，扩散策略（Diffusion Policy）凭借其强大的动作序列生成模型构建能力，成功实现了从高维观测空间到连续动作空间的稳健映射。该策略在处理复杂任务时，展现出了卓越的泛化性能。然而，与基于Transformer架构的方法（如ACT）相比，Diffusion Policy在动作生成精度方面存在明显不足。这一局限性主要源于训练数据多样性的缺乏，以及对多模态信息挖掘与整合的不充分。

As shown in \hyperref[PVE-DP]{Fig.~\ref{PVE-DP}}, to enable the robot to capture richer perceptual data, we propose the \textbf{Proprioceptive-Visual Enhanced Diffusion Policy (PVE-DP)}. Building upon the original Diffusion Policy, this approach incorporates a proprioceptive enhancement module and a visual enhancement module. First, we integrate IMUs at the robot arm’s end-effectors to acquire quaternion data $X_{\text{quat}}(w, x, y, z)$, which satisfies the unit quaternion constraint in Eq.~\eqref{eq:1} and Eq.~\eqref{eq:2}:
\begin{equation}\label{eq:1}
\mathbf{q} = w + x \vec{i} + y \vec{j} + z \vec{k},
\end{equation}
\begin{equation}\label{eq:2}
\left|\mathbf{q}\right|^2 = w^2 + x^2 + y^2 + z^2 = 1.
\end{equation}

In Eq.~\eqref{eq:3}, the end-effector quaternion is incorporated as a new modality and further fused with joint angle data to construct a rich proprioceptive feature representation $P_{\text{fused}}$:
\begin{equation}\label{eq:3}
P_{\text{fused}} = \text{Concat}(X_{\text{qpos}}, X_{\text{quat}}).
\end{equation}

Although forward kinematics can compute end-effector orientation, errors in joint positions inevitably propagate to the estimated quaternion. The IMU is not intended to merely replace FK computation; instead, $X_{\text{quat}}$ provides direct physical orientation feedback, complementing $X_{\text{qpos}}$ at the proprioceptive level.

At the algorithmic level, we introduce a FE-EMA module and deeply integrate it with the visual backbone of the diffusion policy. This module enables efficient fusion of spatial and frequency domain features, thereby constructing spatio-frequency visual feature representations.

\begin{figure*}[htbp]
    \centering
    \includegraphics[width=\textwidth]{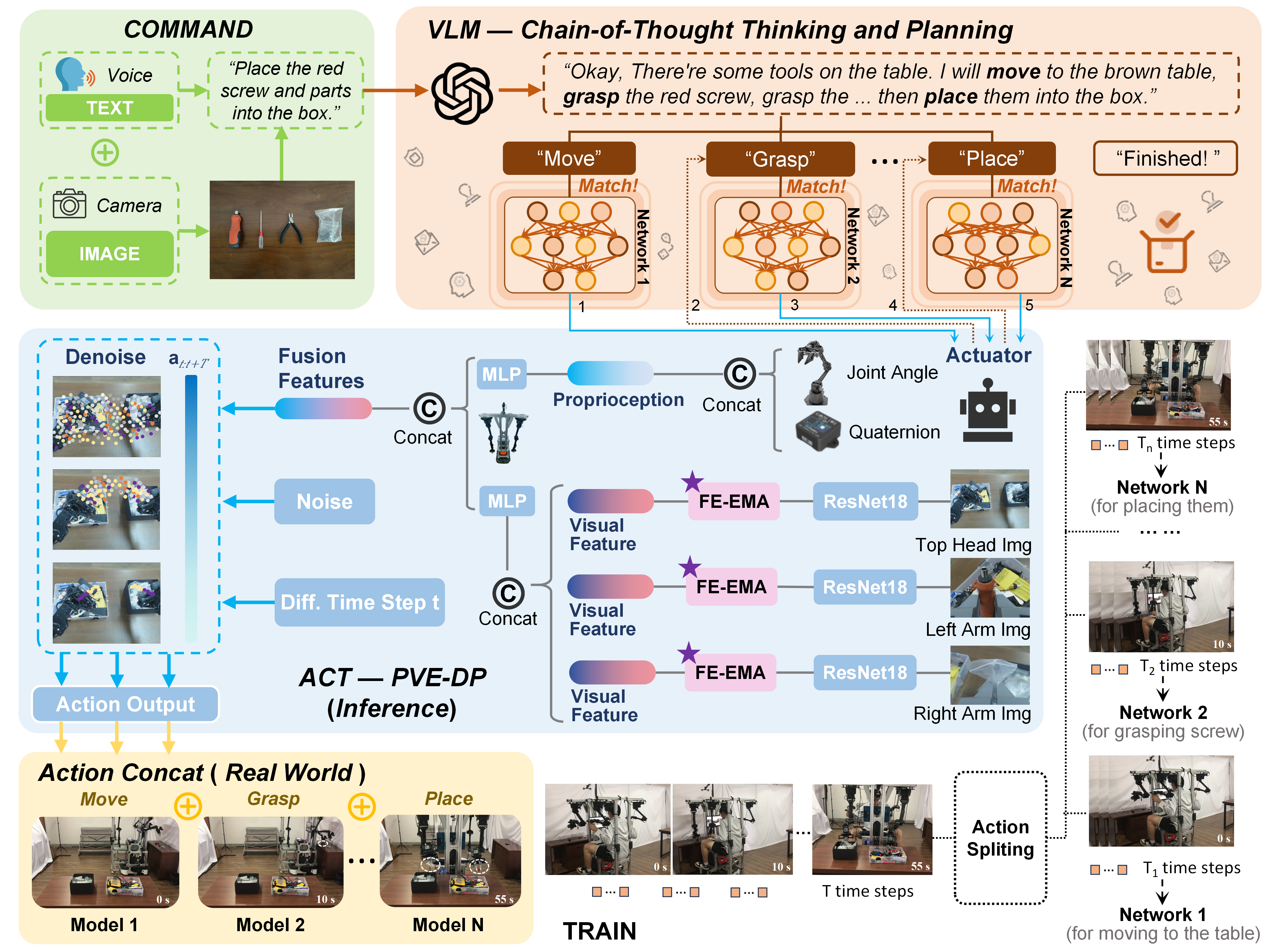} % 适应整个页面宽度
    \captionsetup{singlelinecheck=off, justification=raggedright} % 左对齐标题
    \caption{Overview of AMN framework. This figure presents the Auto-Matching Network architecture, featuring: (1) \textbf{Task Decomposition} of complex task $\mathcal{T}$ into sub-tasks for specialized policy networks; (2) \textbf{Vision-Language Input} processing both linguistic commands and visual scenes; (3) \textbf{Chain-of-Thought Thinking and Planning} that breaks tasks into sequential steps matched with pre-trained policies; (4) \textbf{Inference Execution} using PVE-DP with wrist quaternion and visual spatio-frequency fusion for precise manipulation.}
    \label{AMN}
\end{figure*}

Given an input feature map $X$, its wavelet transform follows Eq.~\eqref{eq:4} via a set of scaled and translated wavelet basis functions $\psi_{j,k}(x)$ and scaling functions $\varphi_{j,k}(x)$:
\begin{equation}\label{eq:4}
X = \sum_{j \in \mathbb{Z}} \sum_{k \in \mathbb{Z}} C_A^{j,k} \cdot \varphi_{j,k}(x) + \sum_{i=1}^{3} C_D^{i,j,k} \cdot \psi_{i,j,k}(x),
\end{equation}
where $j$ is the scale parameter controlling frequency resolution, and $k$ is the translation parameter governing spatial localization. $C_A^{j,k}$ denotes the low-frequency approximation coefficients, while $C_D^{i,j,k}$ represents the high-frequency detail coefficients capturing horizontal, vertical, and diagonal directional details. $\mathbb{Z}$ denotes the set of integers, indicating that $j$ and $k$ can take any integer values.

For a 2D feature map $X$ the wavelet transform can be implemented using 2D filters. It decomposes the feature map along rows and columns, yielding the following coefficients: $\text{DWT}(X)$ = $(cA, cH, cV, cD)$, where $cA$ denotes the low-frequency approximation coefficients, capturing global structural information;$cH$ represents the horizontal detail components, extracting horizontal edge features; $cV$ corresponds to the vertical detail components, capturing vertical edge features; and $cD$ indicates the diagonal detail components, characterizing diagonal edge features.

% The convolution kernel of Haar wavelet is defined in Equation (5), where L is the low-pass filter, which is used for the extraction of low-frequency components. H is a high-pass filter for extracting high-frequency components.

% \begin{equation}
% L = \frac{1}{\sqrt{2}} \begin{bmatrix} 1 \\ 1 \end{bmatrix}, \quad H = \frac{1}{\sqrt{2}} \begin{bmatrix} 1 \\ -1 \end{bmatrix}
% \end{equation}

The high-frequency feature fusion is expressed as $F_H = cH + cV + cD$, combined with the low-frequency feature $cA$ to form an enhanced feature representation $X_{\text{freq}} = \text{ResBlock}(cA + F_H)$. Finally, the FE-EMA module generates a cross-space attention map via matrix dot-product to capture pixel-wise relationships and enhance feature representation. The spatial fusion feature $X_{\text{space}}$ is generated through an attention mechanism and integrated with the frequency-domain feature $X_{\text{freq}}$, as shown in Eq.~\eqref{eq:5}:
% 动态加权参数 $\alpha$ 用于控制时域和频域特征的融合。加权系数由全局平均池化提取的全局特征计算得到，公式为：$\alpha = \sigma(W \cdot GAP(X))$，其中 $\sigma(\cdot)$ 为 Sigmoid 函数，GAP 为全局平均池化操作，$W$ 为权重矩阵。该权重矩阵通过全局平均池化动态优化特征融合过程，允许模型根据任务复杂度动态调整对时域和频域特征的关注程度。
\begin{equation}\label{eq:5}
V_{\text{fused}} = \alpha \cdot X_{\text{space}} + (1 - \alpha) \cdot X_{\text{freq}}.
\end{equation}

In Eq.~\eqref{eq:5}, $X_{\text{space}}$ represents the spatial domain feature generated through an attention mechanism and group normalization process. It primarily captures local regional information extracted from the input features. This feature is combined with the frequency-domain feature $X_{\text{freq}}$ to ensure that the model captures more comprehensive multi-scale information. Such an approach enables dynamic balancing between spatial and frequency domain features.
% , effectively enhancing the model's performance.

% FE - EMA模块基于小波变换理论，结合频域增强技术，能够有效融合时域和频域信息，最终生成具有时域和频域相融合的时 - 频域视觉特征$X_{\text{fused}}$，其过程可表示为
% \begin{equation}
% V_{\text{fused}} = \text{FE - EMA}(\text{ResNet - 18}(X_{\text{image}}))
% \end{equation}

Based on the above multi-modal fusion representation, the loss function of the diffusion policy is defined in Eq.~\eqref{eq:6}:
\begin{equation}\label{eq:6}
\mathcal{L}(\theta) = \mathbb{E}_{t,A_0^k,\epsilon} \left[ \|\epsilon - \epsilon_{\theta}(P_{\text{fused}}, V_{\text{fused}}, A_t^k, t)\|^2 \right],
\end{equation}
where $\theta$ denotes the model parameters, and $\epsilon_{\theta}(P_{\text{fused}}, V_{\text{fused}}, A_t^k, t)$ represents the noise predicted by the model based on the multi-modal fused features $V_{\text{fused}}$ and $P_{\text{fused}}$, noisy action sequence $A_t^k$, and timestep $t$. 

During denoising, PVE-DP continuously refines the predicted action sequence by incorporating additional wrist perception data from robotic arms IMUs, along with spatio-frequency features extracted by the FE-EMA network, enabling precise motion execution in fine manipulation tasks.

\subsection{Auto-Matching Network (AMN) Architecture}

% \begin{figure}[htbp]
%     \centering
%     \includegraphics[width=\linewidth, height=0.5\textheight, keepaspectratio]{Pictures/9_4.png}
%     \captionsetup{singlelinecheck=off, justification=raggedright}
%     \caption{本图展示了在具体任务中DP与PVE-DP的操作细节。}
%     \label{Experient_PVE-DP}
% \end{figure}

A critical challenge in robot imitation learning lies in long-horizon complex tasks. As execution extends, inference errors accumulate, while the absence of logical dependencies between action segments hinders a single policy from generalizing to multi-skill, multi-scenario applications.

As shown in \hyperref[AMN]{Fig.~\ref{AMN}}, we propose the \textbf{Auto-Matching Network (AMN)} framework. This architecture integrates the vision-language model GLM-4.1V~\cite{RN55} and employs a ``Chain of Thought'' reasoning paradigm to decompose complex long-horizon tasks into logically connected subtask sequences. It dynamically matches each subtask with a pre-trained policy network for distributed inference. The core modeling concept can be expressed as Eq.~\eqref{eq:7}:
\begin{equation}\label{eq:7}
\mathcal{T} = \{T_i\}_{i=1}^n, \quad T_i \sim \pi_i(s_t,a_t|\theta_i).
\end{equation}

The architecture leverages the Chain of Thought cognitive paradigm to decompose a complex long-horizon task  $\mathcal{T}$ into a sequence of logically connected subtasks $\{T_i\}$. During training, for tasks spanning over 3000+ time steps, we introduce a step-wise saving mechanism: upon completion of each subtask $T_i$, the current state $s_t$ and action $a_t$ are saved as a subtask sample. Through this temporal decoupling strategy, the original task is decomposed into $n$ short-horizon subtasks $\{T_1, \ldots, T_n\}$, the $i$-th of which is trained with an independent network architecture $\pi_i(\cdot|\theta_i)$. Due to the significant differences in perception-action patterns across subtasks, the learned network weights for each subtask exhibit high specificity.

% During inference, the operator provides a command via voice or text. The robot observes the scene and sends the image and instruction to the VLM. The AMN architecture decomposes the task into sequential steps, and then assigns each step a pre-trained specialized policy network for execution. The PVE-DP policy is matched to perform robotic manipulation. After each subtask, the system automatically selects the next network until the entire task is completed.

During inference, the operator provides a command via voice or text. The robot observes the scene and sends the image and instruction to the VLM. The AMN architecture decomposes the task into sequential steps grounded in the pre-trained policy library, and then assigns each step a pre-trained specialized policy network for execution. After each subtask, the system automatically transitions to the next network until the entire task is completed.

\section{EXPERIMENT AND ANALYSIS}

This study conducts experimental validation based on the MuJoCo simulation platform and the RoboMatch real-world robot platform from Table~\ref{tab:data}. On both platforms, we designed and carried out the following three verification tasks: (1) Stability of the AMN architecture in long-horizon reasoning tasks; (2) Fine manipulation capability of the PVE-DP policy in simulated and real-world tasks; (3) Statistical analysis of data collection completion rate and efficiency to evaluate RoboMatch's applicability.

\begin{table}[htbp]
\centering
\caption{We train PVE-DP with short demonstrations from both simulated and real environments per task, while AMN is trained with long real-world demonstrations for each task.}
\label{tab:data} % 这是重要的标签
% \resizebox{\linewidth}{!}{
\begin{tabular}{c c c c c c}
\toprule
\textbf{Scene} & \textbf{No.} & \textbf{Task} & \textbf{Mode}  & \textbf{Make-Up}  & \textbf{Length} \\
\midrule
\multirow{2}{*}{\textbf{Sim}} & 1 & Transfer Cube & Short & 100 Demos & 0.2 Hrs \\
 & 2 & Insertion Scripted & Short & 100 Demos & 0.2 Hrs \\
\midrule
\multirow{4}{*}{\textbf{Real}} 
 & 3 & Wipe Table & Short & 50 Demos & 0.3 Hrs \\
 & 4 & Sort Workpiece & Short & 50 Demos & 0.4 Hrs \\
 & 5 & Clean Trash & Long & 50 Demos & 1.1 Hrs \\
 & 6 & Deliver Tool & Long & 50 Demos & 1.4 Hrs \\
\bottomrule
\end{tabular}
% }
\end{table}

\subsection{Experiments of AMN Architecture}
As mentioned above, the AMN framework generates task plans based on the scene and instructions, decomposes them into sub-tasks, and matches each to specialized networks for long-horizon task completion. To evaluate AMN's performance in long-horizon tasks, we compare it against baseline models including ACT, DP, and PVE-DP. Two long-horizon tasks are designed: a four-step task, \textit{Clean Trash} (3000 time steps), and a six-step task, \textit{Deliver Tool} (4000 time steps).

\begin{figure*}[htbp]
    \centering
    \includegraphics[width=\textwidth]{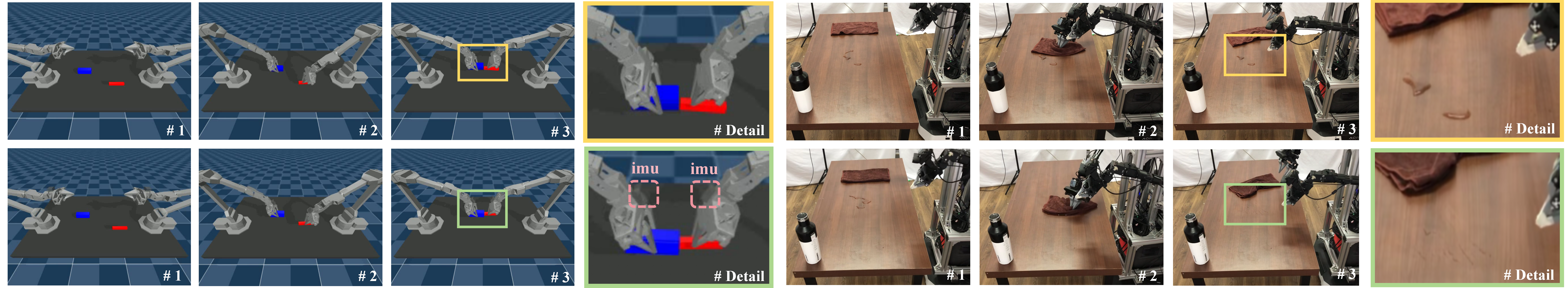} % 适应整个页面宽度
    \captionsetup{singlelinecheck=off, justification=raggedright}
    \caption{Comparison between DP and the proposed PVE-DP method in simulation and real-world tasks. The first row demonstrates the inference results of DP, while the second row presents the improved performance of PVE-DP.}
    \label{Experient_PVE-DP}
\end{figure*}

\begin{table*}[t]
\centering
\caption{Long-horizon inference performance comparison across different methods and tasks (50 trials total). Underlined values indicate the second-highest success rates for the current task, while bolded values represent the highest success rates.}
\label{tab:results}
% \resizebox{\textwidth}{!}{%
\begin{tabular}{ccccccccccc}
\toprule
\multirow{3}{*}{Method} & \multicolumn{4}{c}{\textbf{Clean Trash (3000 time steps)}} & \multicolumn{6}{c}{\textbf{Deliver Tool (4000 time steps)}} \\
\cmidrule(lr){2-5} \cmidrule(lr){6-11}
 & Grasp & Move & Throw & Back & Move & \begin{tabular}{@{}c@{}}Grasp\\(screw)\end{tabular} & \begin{tabular}{@{}c@{}}Grasp\\(pouch)\end{tabular} & Move & Put & Back \\
\midrule
ACT~\cite{RN4} & 54\% & \underline{46\%} & \underline{26\%} & \underline{10\%} & 100\% & 52\% & \underline{34\%} & \underline{24\%} & \underline{6\%} & \underline{2\%} \\
DP U-Net~\cite{RN5} & 30\% & 14\% & 0\% & 0\% & 100\% & 26\% & 8\% & 0\% & 0\% & 0\% \\

PVE-DP (ours) & \underline{66\%} & 42\% & 20\% & 8\% & 100\% & \underline{54\%} & 28\% & 8\% & 0\% & 0\% \\
\textbf{AMN (Ours)} & \textbf{74\%} & \textbf{70\%} & \textbf{54\%} & \textbf{50\%} & 100\% & \textbf{68\%} & \textbf{66\%} & \textbf{62\%} & \textbf{48\%} & \textbf{44\%} \\
\bottomrule
\end{tabular}%
% }
\end{table*}

% \begin{figure}[htbp]
%     \centering
%     \includegraphics[width=\columnwidth]{Pictures/7.png} % 适应单栏宽度
%     \captionsetup{singlelinecheck=off, justification=raggedright}
%     \caption{该图展示了Split-Type和RoboMatch的操作形式。}
%     \label{Split-Type_And_RoboMatch}
% \end{figure}

In terms of training strategy, AMN employs a phased training approach where individual subtask modules are trained independently and temporally composed during inference. In contrast, baseline methods adopt an end-to-end training strategy that directly feeds complete sequential data into the network.

As shown in Table~\ref{tab:results}, AMN maintains a more stable success rate during long-horizon inference, outperforming other policies by over 40\% in final task completion. While ACT exceeds DP and PVE-DP in long-term inference, its full-task success remains below 10\%, further highlighting AMN's superior capability in complex scenarios.

\subsection{Ablation Experiments of PVE-DP}

To thoroughly evaluate the performance of PVE-DP, we compare it against state-of-the-art baseline models in the field. The experiments include four tasks: two simulated tasks from the ALOHA benchmark (cube transfer and bimanual insertion), and two real-robot tasks (wiping table and sorting workpiece). To facilitate algorithm validation, IMU modules were integrated into both the simulation platform and the physical RoboMatch test platform.

% For each task, target objects are randomly placed within a predefined range. 

\begin{table}[h]
\centering
\caption{Comparison of Parameters and FLOPs.}
\label{tab:comparison}
\begin{tabular}{clcc}
\toprule
\multirow{2}{*}{Method} & \multirow{2}{*}{Backbone} & \multicolumn{2}{c}{Simulation Tasks} \\
\cmidrule{3-4}
 & & \#.Param. (M) & FLOPs (B) \\
\midrule
Diffusion Policy (DP) & \multirow{3}{*}{ResNet18} & 101.7214 & 33.9266 \\
DP + FE-EMA & & 101.7269 & 33.9662 \\
DP + FE-EMA + Quat & & \textbf{101.8416} & \textbf{33.9664} \\
\bottomrule
\end{tabular}
\end{table}

\begin{table}[h]
\centering
\caption{\small Comparison of Success Rate (\%) for Simulation and Real-World Tasks.}
\label{tab:ablation_icra}
\renewcommand{\arraystretch}{1.4}
\begin{tabular}{ccccccc}
\toprule
Method & \cellcolor{gray!20}Avg. & Task1 & Task2 & Task3 & Task4 \\
\midrule
ACT~\cite{RN4} & \cellcolor{gray!20}52.0 & 86 & 32 & 48 & 42 \\
DP U-Net~\cite{RN5} & \cellcolor{gray!20}43.8 & 80 & 36 & 39 & 20 \\
NL-ACT~\cite{RN9} & \cellcolor{gray!20}53.8 & 86 & 40 & 43 & 46 \\
InterACT~\cite{RN10} & \cellcolor{gray!20}58.8 & 88 & 46 & 51 & 50 \\
DiT-Block~\cite{RN3} & \cellcolor{gray!20}56.8 & 89 & 56 & 47 & 35 \\
DP w. EMA~\cite{RN43} & \cellcolor{gray!20}54.0 & 90 & 42 & 43 & 41 \\
\midrule
ACT w. quat. (ours) & \cellcolor{gray!20}63.0 & 96 & 56 & 54 & 46 \\
DP w. quat. (ours) & \cellcolor{gray!20}61.0 & 98 & 52 & 51 & 43 \\
DP w. FE-EMA (ours) & \cellcolor{gray!20}59.0 & 94 & 48 & 47 & 47 \\
\textbf{PVE-DP (ours)} & \cellcolor{gray!20}\textbf{75.3} & \textbf{100} & \textbf{79} & \textbf{63} & \textbf{59} \\
\bottomrule
\end{tabular}
\end{table}

\begin{figure}[htbp]
    \centering
    \includegraphics[width=\columnwidth]{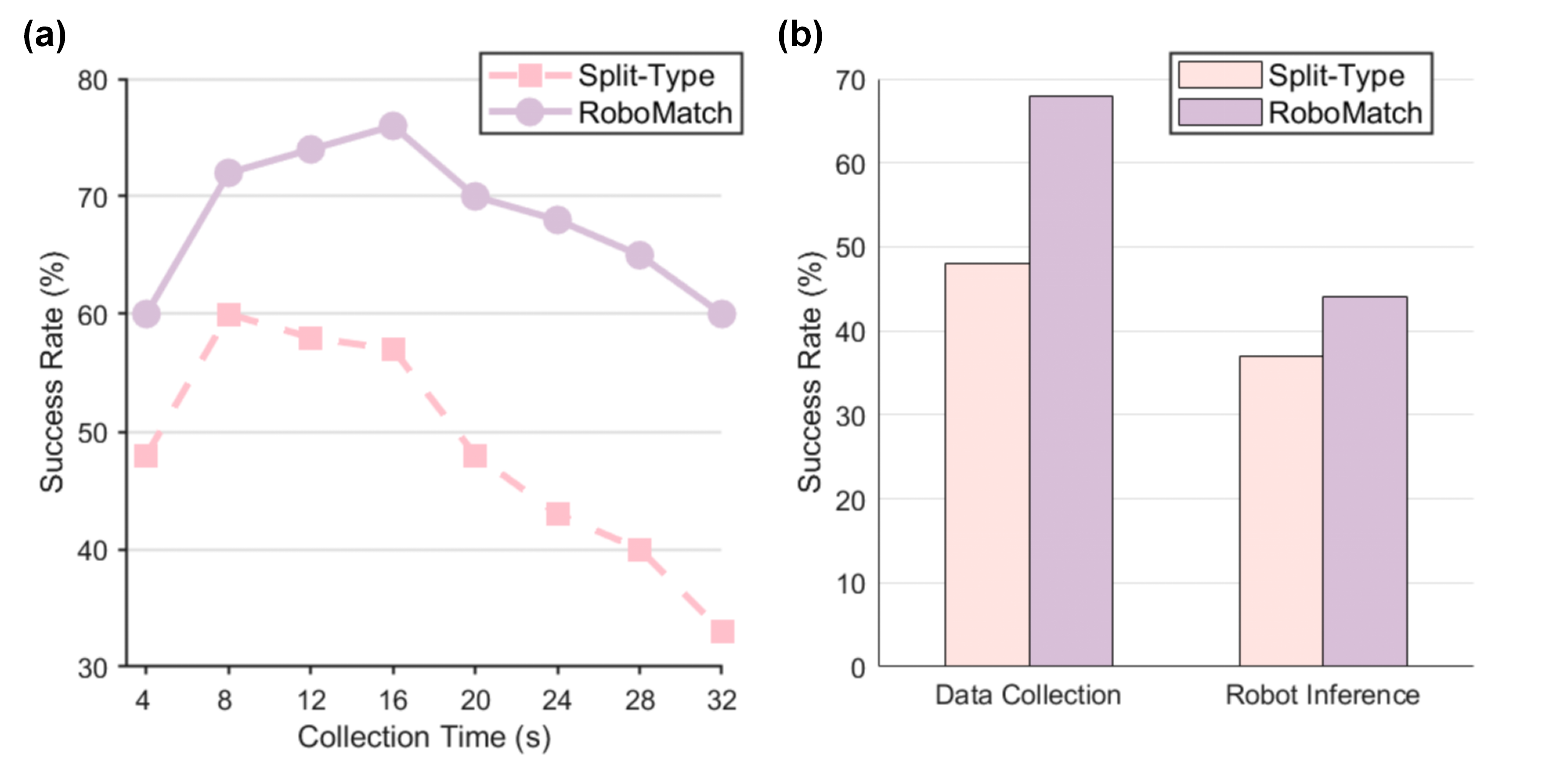} % 适应单栏宽度
    \captionsetup{singlelinecheck=off, justification=raggedright}
    \caption{The impact of RoboMatch on performance. (a) Success rate vs. task duration: Split-Type vs. RoboMatch during data collection. (b) Average success rate: Split-Type vs. RoboMatch across both data collection and inference.}
    \label{Experient_RoboMatch}
\end{figure}

As shown in Table~\ref{tab:comparison} and Table~\ref{tab:ablation_icra}, our method achieves an average success rate of 75.3\% without significantly increasing the number of parameters or FLOPs, outperforming all baseline methods across all four tasks. Combined with \hyperref[Experient_PVE-DP]{Fig.~\ref{Experient_PVE-DP}}, these results demonstrate the superiority of the proposed approach in manipulation tasks. Furthermore, our method consistently exceeds baseline performance with both scripted and human demonstrations, indicating its strong capability to capture the multimodal characteristics of human data.

\subsection{RoboMatch Performance Experiment}

% 为充分评估RoboMatch平台的性能，我们选取分体式数据采集平台作为对比，定义了两组对照实验，对比数据采集效率和数据采集的质量。我们针对Table ~\ref{tab:Data}中的Real任务和其他自定义任务(共计100 demos)进行数据采集，这些任务涵盖4s到32s，间隔为4s的采集任务。我们记录下分体式平台和RoboMatch平台一次性完成数据任务的次数,并将采集的数据放入同一Policy中进行训练和推理,测试不同方式采集的数据对训练效果的影响。
% As shown in \hyperref[Experient_RoboMatch]{Fig.~\ref{Experient_RoboMatch}}，随着数据采集任务的时间增长，RoboMatch在采集数据时的成功率整体显著高于分体式平台，且随着时间增长，平均成功率方面较分体式平台平均提升约 25\%,体现了其对采集数据能力显著的提升。与此同时，RoboMatch采集的数据放入机器人策略训练后，推理时相较于分体式成功率提升约7\%,这体现了一体式高质量数据对机器人自主推理起到了有效的提升。
To evaluate RoboMatch systematically, this study uses a split-type (locomotion and manipulation) data collection platform for baseline comparison through a dual-metric (efficiency and quality) analysis. The test set includes 100 demonstrations—containing real tasks from Table~\ref{tab:data} and custom tasks—with duration varying from 4s to 32s at 4s intervals. We measure the one-time collection success rate for both platforms and train a unified policy on the obtained data to evaluate quality differences via inference testing.

As shown in \hyperref[Experient_RoboMatch]{Fig.~\ref{Experient_RoboMatch}}, RoboMatch exhibits significantly higher collection stability with increasing task duration, consistently outperforming the decoupled platform. Quantitative results show a 20\% average improvement in success rate and a 7\% higher inference success rate for policies trained with RoboMatch data, demonstrating its effectiveness in enhancing robotic autonomy.

\section{CONCLUSIONS}

% A conclusion section is not required. Although a conclusion may review the main points of the paper, do not replicate the abstract as the conclusion. A conclusion might elaborate on the importance of the work or suggest applications and extensions. 
In this paper, we propose RoboMatch, a collaborative, precise, portable, and flexible platform designed for unified whole-body mobile-manipulation teleoperation, which enables the accurate execution of operational tasks and supports long-horizon inference in real-world environments. The proposed platform significantly reduces the complexity and human effort of teleoperation. Additionally, the PVE-DP method enhances fine manipulation capabilities, while the AMN framework improves operational flexibility, enabling the seamless integration of diverse skills and increasing success rates in long-horizon tasks. Future work will focus on optimizing the RoboMatch cockpit to enhance operator ergonomics during data collection, refining the PVE-DP method to further elevate robotic manipulation performance, and smoothing the transitions between subtasks within the AMN framework to facilitate more cohesive inference.

\bibliographystyle{IEEEtran}%规定参考文献的样式
% \bibliography{refs}  %参考文献库的名字Refs

\end{document}